\documentclass{article}

\usepackage{PRIMEarxiv}

\usepackage[utf8]{inputenc} 
\usepackage[T1]{fontenc}    
\usepackage{hyperref}       
\usepackage{url}            
\usepackage{booktabs}       
\usepackage{amsfonts}       
\usepackage{nicefrac}       
\usepackage{microtype}      
\usepackage{lipsum}
\usepackage{fancyhdr}       
\usepackage{graphicx}       
\graphicspath{{media/}}     

\usepackage{multirow}
\usepackage{amsmath}
\usepackage{amssymb}
\usepackage{amsthm}
\usepackage{mathrsfs}
\usepackage[title]{appendix}
\usepackage{xcolor}
\usepackage{textcomp}
\usepackage{manyfoot}
\usepackage{booktabs}
\usepackage{algorithm}
\usepackage{algorithmicx}
\usepackage{algpseudocode}
\usepackage{listings}

\newcommand{\methodname}{CHASMA}
\newcommand{\twitterdataset}{VMU-Twitter}
\newcommand{\AMM}{Asymmetric-MM}
\newcommand{\MMD}{MMD}
\newcommand{\benchmark}{VERITE}

\title{VERITE: A Robust Benchmark for Multimodal Misinformation Detection Accounting for Unimodal Bias}

\usepackage{authblk}

\author[1, 2]{\small Stefanos-Iordanis Papadopoulos\thanks{Corresponding author} \ }
\author[1]{\small Christos Koutlis}
\author[1]{\small Symeon Papadopoulos}
\author[2]{\small Panagiotis C. Petrantonakis}

\affil[1]{\footnotesize Information Technology Institute, Centre for Research \& Technology, Hellas.}
\affil[2]{\footnotesize Department of Electrical \& Computer Engineering, Aristotle University of Thessaloniki.}
\affil[ ]{\textit {\{stefpapad,ckoutlis,papadop\}@iti.gr, \textit{ppetrant@ece.auth.gr}}}

\begin{document}
\maketitle

\begin{abstract}Multimedia content has become ubiquitous on social media platforms, leading to the rise of multimodal misinformation (MM) and the urgent need for effective strategies to detect and prevent its spread.
In recent years, the challenge of multimodal misinformation detection (\MMD) has garnered significant attention by researchers and has mainly involved the creation of annotated, weakly annotated, or synthetically generated training datasets, along with the development of various deep learning \MMD \ models.
However, the problem of unimodal bias has been overlooked, where specific patterns and biases in \MMD \ benchmarks can result in biased or unimodal models outperforming their multimodal counterparts on an inherently multimodal task; making it difficult to assess progress. 
In this study, we systematically investigate and identify the presence of unimodal bias in widely-used \MMD \ benchmarks, namely \twitterdataset \ and COSMOS. 
To address this issue, we introduce the ``VERification of Image-TExt pairs'' (\benchmark) benchmark for \MMD \ which incorporates real-world data, excludes ``asymmetric multimodal misinformation'' and utilizes ``modality balancing''.
We conduct an extensive comparative study with a Transformer-based architecture that shows the ability of \benchmark \ to effectively address unimodal bias, rendering it a robust evaluation framework for \MMD.  
Furthermore, we introduce a new method -termed Crossmodal HArd Synthetic MisAlignment (\methodname)- for generating realistic synthetic training data that preserve crossmodal relations between legitimate images and false human-written captions.
By leveraging \methodname \ in the training process, we observe consistent and notable improvements in predictive performance on \benchmark; with a 9.2\% increase in accuracy.
We release our code at: \url{https://github.com/stevejpapad/image-text-verification}

\end{abstract}


\section{Introduction}

\begin{figure*}[!t]
\centering
\includegraphics[width=6in]{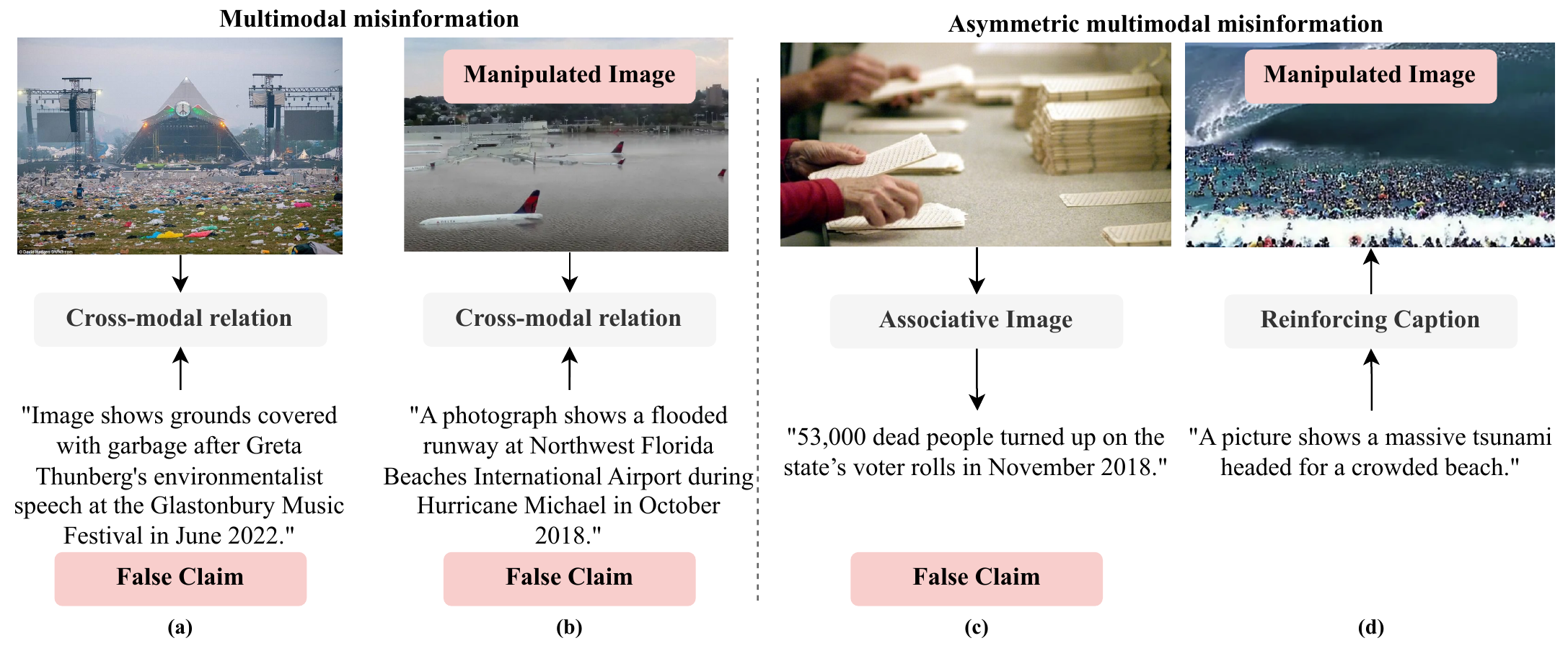}
\caption{Examples of multimodal misinformation (taken from \benchmark) and asymmetric multimodal misinformation (taken from COSMOS benchmark).}
\label{fig:amm_vs_mm}
\end{figure*}

The proliferation of misinformation poses a significant societal challenge with potential negative impacts on democratic processes \cite{bennett2018disinformation}, social cohesion \cite{duffy2020too}, public health \cite{roozenbeek2020susceptibility}, political and religious persecution \cite{gamir2021multimodal} among others.
The widespread usage of digital media platforms in recent years has only exacerbated the problem \cite{olan2022fake}. 
In the context of social media platforms, multimedia content has been shown to often be more attention-grabbing and widely disseminated than plain text \cite{li2020picture}, while the presence of an image can significantly enhance the persuasiveness of a false statement \cite{newman2012nonprobative}.
Against this backdrop, while the work of fact-checkers becomes increasingly important it also becomes increasingly more difficult, considering the scale of content produced and shared daily on social media. 
In response, researchers have been investigating a range of AI-based methods for detecting misinformation, e.g. detecting inaccurate claims with the use of natural language processing \cite{mridha2021comprehensive}, detecting synthetic images, such as DeepFakes, with the use of deep learning \cite{rana2022deepfake} or multimodal misinformation with the use of multimodal deep learning \cite{hangloo2022combating}. 

Multimodal misinformation (MM) typically refers to false or misleading information that is spread using multiple modes of communication, such as text, images, audio and video \cite{alam2021survey}. 
Here, we focus on image-caption pairs that collaboratively contribute to the dissemination of misinformation.
For instance, in Fig. \ref{fig:amm_vs_mm}a, an image depicts the grounds of a musical festival covered in garbage, accompanied by the false claim that it was taken in June 2022 ``after Greta Thunberg's environmentalist speech'', while the image was actually taken in 2015\footnote{\url{https://www.snopes.com/fact-check/glastonbury-greta-thunberg}}.

Previous studies on automated multimodal misinformation detection (\MMD) have predominantly explored three approaches in terms of training datasets: annotated (\cite{boididou2018verifying, zlatkova2019fact}), weakly-annotated (\cite{nielsen2022mumin, jindal2020newsbag, nakamura2019r}) and synthetically generated datasets \cite{jaiswal2017multimedia, luo2021newsclippings, sabir2018deep}. 
These distinct routes facilitated the development and evaluation of multimodal models designed to detect and combat misinformation effectively \cite{wang2018eann, khattar2019mvae, singhal2019spotfake, yu2022bcmf}.
However, previous studies have overlooked the investigation of unimodal bias. Training datasets exhibiting certain patterns and biases (asymmetries and imbalances) towards one modality can lead to biased models or unimodal methods capable of outperforming their multimodal counterparts in a purportedly multimodal task.
If these patterns persist within the evaluation benchmarks, they can obscure the impact of unimodal bias; hindering our ability to effectively assess progress in the field of \MMD.
In our investigation, we uncover that the widely used \twitterdataset \ dataset \cite{boididou2018verifying}
exhibits an image-side unimodal bias 
while the COSMOS evaluation benchmark \cite{aneja2021cosmos}
exhibits a text-side unimodal bias; 
raising questions about their reliability as evaluation benchmarks for \MMD.

Against this backdrop, the primary aim of this study is to create a robust evaluation framework that accounts for unimodal bias. 
To this end, we have create the ``VERification of Image-TExt pairs'' (\benchmark) evaluation benchmark which accounts for unimodal bias by (1) consisting of real-world data, (2) excluding ``asymmetric multimodal misinformation'' (\AMM) and (3) employing ``modality balancing''. 
We introduce the term \AMM \ -and contrast it with MM- to highlight cases where one dominant modality is responsible for propagating misinformation while other modalities have little or no influence.
An example of \AMM \ can be seen in Fig. \ref{fig:amm_vs_mm}c where a claim pertains to ``deceased people turning up to vote'' and an image merely thematically related to the claim, is added primarily for cosmetic enhancement.
Focusing on the dominant modality, a robust text-only  (or, in other scenarios, image-only) detector would suffice for detecting misinformation; rendering the other modality inconsequential in the detection process. 
We hypothesize that this asymmetry can exacerbate unimodal bias.
Furthermore, we introduce the concept of ``modality balancing'' which ensures that all images and captions are presented twice during evaluation, once in their truthful pair and once in their misleading pair,
thus compelling a model to consider both modalities and their relation when discerning between truth and misinformation.
We conduct a comprehensive comparative analysis where we train a Transformer-based architecture using different datasets, including \twitterdataset, Fakeddit, and various synthetically generated datasets. 
Our empirical results demonstrate that \benchmark \ effectively mitigates and prevents the occurrence of unimodal bias. 

Our second contribution is the introduction of ``Crossmodal HArd Synthetic MisAlignment'' (\textit{\methodname}), a new method for generating synthetic training datasets that aims to maintain crossmodal relation between legitimate images and misleading human-written texts to create plausible misleading pairs.  
More specifically, \textit{\methodname} utilizes a large pre-trained crossmodal alignment model (CLIP \cite{radford2021learning}) to pair legitimate images (from VisualNews \cite{liu2020visual}) with contextually relevant but misleading texts (from Fakeddit). 
\textit{\methodname} maintains the sophisticated linguistic patterns (e.g. exaggeration, irony, emotions) that are often found in human-written texts; unlike methods that rely on Named Entity Inconsistencies (NEI) for generating MM \cite{sabir2018deep}.
The inclusion of \textit{\methodname} in the training process consistently enhances the predictive performance on the \benchmark \ benchmark, particularly evident in aggregated datasets, resulting in a notable 9.2\% increase in accuracy.

The main contributions of our work can be summarised as follows:
\begin{itemize}

\item Systematically investigate the issue of unimodal bias within widely used evaluation \MMD \ benchmarks (\twitterdataset \ and COSMOS). 

\item Create the \benchmark \ benchmark, which effectively mitigates the problem of unimodal bias and provides a more robust and reliable evaluation framework for \MMD.

\item Develop \methodname, a novel approach for creating synthetic training data for \MMD, that consistently leads to improved detection accuracy on the \benchmark \ benchmark.

\end{itemize}

\section{Related Work}
\label{sec:related_work}

The automated detection of misinformation is a challenging task that has garnered increasing attention from researchers in recent years. 
A range of methods is being explored to identify misinformation in text \cite{mridha2021comprehensive} and images \cite{rana2022deepfake}.
Consequently, multiple datasets have been created for fake news detection \cite{thorne2018fever} 
and manipulated images \cite{heller2018ps}. These challenges involve unimodal settings. However, there is a need for \MMD \ models that can handle cases where the combination of an image and its caption lead to misinformation. 
Given the complexity of this task, large training datasets are required to train robust \MMD \ models. 
In this section, we explore the available research on existing datasets, including both annotated and synthetically generated as well as available evaluation benchmarks for \MMD.

\subsection{Annotated multimodal misinformation datasets}

The ``image verification corpus'', often referred as the ``Twitter'' dataset, (``\twitterdataset'' from now on) was used in the MediaEval2016 Verifying Multimedia Use (VMU) challenge \cite{boididou2018verifying} and comprises 16,440 tweets regarding 410 images for training and 1,090 tweets regarding 104 images for evaluation. 
Since the images are accompanied by tweets, the dataset has been widely used for \MMD \ \cite{wang2018eann, khattar2019mvae, singhal2019spotfake, singhal2019spotfake, yu2022bcmf}.
In addition, the Fauxtography dataset comprises manually fact-checked image-caption pairs 
sourced from Snopes\footnote{\url{https://www.snopes.com/fact-check/category/photos}} and Reuters\footnote{\url{https://www.reuters.com/fact-check}}, with a total of 1,233 pairs, of which 592 are classified as truthful and 641 as misleading \cite{zlatkova2019fact}. 
However, their very limited size raises doubts about the effectiveness and generalizability of deep neural networks trained on these datasets. 

To address the challenges of collecting and annotating large-scale datasets, researchers have also explored weakly annotated datasets. 
The MuMiN dataset, for instance, consists of 21 million tweets on Twitter, linked to 13,000 fact-checked claims, with a total of 6,573 images \cite{nielsen2022mumin}. 
While this dataset provides rich social information such as user information, articles, and hashtags, its limited number of images may also be insufficient for \MMD.
The NewsBag is another large-scale multimodal dataset that was created by scraping the Wall Street Journal and Real News for truthful pairs and The Onion\footnote{\url{https://www.theonion.com}} and The Poke\footnote{\url{https://www.thepoke.co.uk}} for misleading pairs \cite{jindal2020newsbag}. However, the latter sites publish humorous and satirical articles which may not reflect real-world misinformation \cite{levi2019identifying}. 

Fakeddit is a large weakly labeled dataset consisting of 1,063,106 instances collected from various subreddits\footnote{\url{https://www.reddit.com}} and grouped into two, three, or six classes based on their content \cite{nakamura2019r}. 
The instances are classified as either Truthful or Misleading and then separated into six classes, including true, satire, misleading content, manipulated content, false connection, or impostor content. Of the total instances, 680,798 have both an image and a caption, with 413,197 of them being Misleading and 267,601 being Truthful.
Despite being weakly labeled, Fakeddit provides a large-scale resource for training machine learning models to detect misleading multimodal content.

\subsection{Synthetic multimodal misinformation datasets}
\label{sec:rw_misinformers}

Due to the need for large-scale datasets, the labor-intensive nature of manual annotation and the potential for weak labeling to introduce noise, researchers have also been exploring the use of synthetically generated training data for \MMD. 
These methods can be categorized into two groups based on the type of misinformation they generate, namely OOC pairs or NEI. 

OOC-based datasets can be created through random-sampling techniques, such as in the case of the MAIM \cite{jaiswal2017multimedia} and COSMOS \cite{aneja2021cosmos} datasets. 
However, these methods tend to produce easily detectable non-realistic pairs, making them unsuitable for training effective misinformation detection models \cite{papadopoulos2023synthetic}.
An alternative approach is to use feature-based sampling to retrieve more realistic pairs that more realistically resemble multimodal misinformation.
The NewsCLIPings dataset \cite{luo2021newsclippings} was created using scene-learning, person matching and CLIP in order to retrieve images from within the VisualNews dataset in order to create OOC samples.
Similarly, the Twitter-COMMs dataset was created via CLIP-based sampling on Twitter data related to climate, COVID, and military vehicles \cite{biamby2021twitter}.

On the other hand, NEI-based methods rely on substituting named entities in the caption - such as people, locations, and dates - with other entities of the same type, resulting in misleading inconsistencies between the image and caption.
Since random retrieval and replacement of entities may be easily detectable \cite{papadopoulos2023synthetic}, several methods have been proposed to retrieve relevant entities based on cluster-based retrieval for MEIR \cite{sabir2018deep}, rule-based retrieval for TamperedNews \cite{muller2020multimodal}, and CLIP-based retrieval for CLIP-NESt \cite{papadopoulos2023synthetic}. 
Finally, aggregating synthetically generated datasets - combining both OOC and NEI - has been shown to further improve performance \cite{papadopoulos2023synthetic}. 

\subsection{Unimodal Bias and Evaluation Benchmarks}
\label{sec:eval_n_bias}

Unimodal bias has mainly been observed and investigated in the domain of visual question answering (VQA), wherein biased models rely on surface-level statistical patterns within one modality (usually the textual modality) while disregarding the information present in the other modality (usually the visual modality) \cite{goyal2017making}. 
Evaluation benchmarks have been devised to enhance fairness and robustness of evaluating of VQA models \cite{agrawal2018don} and various methods have been proposed for counteracting unimodal bias during training \cite{cadene2019rubi}.
However, comparable efforts in addressing unimodal bias have not been explored within the context of \MMD.

Currently, there is no widely accepted benchmark for evaluating \MMD \ models. 
Most studies assess their approaches on a split of their weakly annotated \cite{nakamura2019r, nielsen2022mumin} or their synthetically generated datasets \cite{jaiswal2017multimedia, luo2021newsclippings,  sabir2018deep, muller2020multimodal}, 
which may not provide a realistic estimate of how these methods will perform when confronted with real-world misinformation. 
The COSMOS benchmark is one of the few works that collect an evaluation set consisting of real-world multimodal misinformation and make it publicly available \cite{aneja2021cosmos}. 
It consists of 1,700 pairs and is balanced between truthful and misleading pairs -collected from credible news sources and Snopes.com respectively- and has been used in two challenges for ``CheapFakes detection'' \cite{aneja2021mmsys, aneja2022acm}.
Nevertheless, in \cite{papadopoulos2023synthetic} it was found that text-only methods, especially NEI-based ones, can outperform their multimodal counterpart on COSMOS, raising questions about its reliability as an \MMD \ benchmark.
Another widely used dataset for \MMD \ is the \twitterdataset \ dataset \cite{boididou2018verifying},
despite consisting mainly of manipulated and digitally created images.
In this paper, we systematically investigate the factors behind unimodal bias in \MMD \ and create a new evaluation benchmark that accounts for it. 

\section{Methodological Framework}

\subsection{Problem Definition}
\label{sec:problem}

In this study, we focus on the challenge of multimodal misinformation detection (\MMD). 
and specifically on image-caption pairs that collaboratively contribute to the propagation of misinformation. 
Typically, \MMD \ can be defined as follows: given a dataset ${(x_i, y_i)}_{i=1}^{N}$, where $x_i = (I_i, C_i)$ represents an image-caption pair and $y_i \in \{0, 1\}$ denotes the ground truth label indicating the presence or absence of misinformation, the objective is to learn a mapping function $f: x \rightarrow y$ that accurately predicts the presence of misinformation in a given image-caption pair.
However, instead of addressing \MMD \ as a binary classification problem (\cite{
zlatkova2019fact, aneja2021cosmos, papadopoulos2023synthetic, 
jaiswal2017multimedia, luo2021newsclippings, sabir2018deep, muller2020multimodal}), we introduce a new taxonomy that includes three classes: 
\begin{enumerate}
    \item Truthful (True): an image-caption pair ($I_i^t, C_i^t$) is considered True when the origin, content, and context of an image are accurately described in the accompanying caption.
    \item Out-Of-Context (OOC) image-text pairs: involves a deceptive combination of a truthful caption $C_i^t$ and an out-of-context image $I_i^x$ or a legitimate image $I_i^t$ with an out-of-context caption $C_i^x$; with ``$^x$'' denoting the different context but otherwise truthful information.  
    \item MisCaptioned images (MC): involves an image $I_i^t$ being paired with a misleading caption $C_i^f$ that misrepresents the origin, content, and/or meaning of the image; with ``$^f$'' denoting falsehood or manipulation.  
\end{enumerate}
We consider the structural differences between OOC and MC to warrant separate classification since MC cases predominantly involve the introduction of falsehoods within the textual modality that are linked to the image, whereas OOC scenarios involve the juxtaposition of otherwise truthful text with a legitimate yet decontextualized image, resulting in the propagation of misinformation.

Furthermore, we investigate the problem of unimodal bias in the context of \MMD, the phenomenon of unimodal models or models biased towards one modality outperforming their unbiased multimodal counterparts on an inherently multimodal task.
Unimodal bias can emerge during the training process as a consequence of certain patterns and biases, wherein models tend to emphasize superficial statistical correlations within a single modality. 
If these patterns persist within the evaluation benchmarks, they have the potential to obscure the presence of unimodal biases within the results.
We hypothesize that one such problematic pattern is ``asymmetric multimodal misinformation'' (\AMM) -which we contrast against MM - where false claims are accompanied by a loosely connected image (associative imagery) or manipulated images are accompanied by captions that simply reinforce the misleading content of the image (reinforcing captions). 
Examples are provided in Figures \ref{fig:amm_vs_mm}c and \ref{fig:amm_vs_mm}d.
Both scenarios create an asymmetry between the two modalities; rendering one modality as the dominant source of misinformation while the second modality has little or no influence. 
It is important to note that instances of MC images (including NEI) may exhibit a certain degree of `asymmetry' in that misinformation is primarily propagated through the textual modality. Nevertheless, we do not consider them to be \AMM \ because the text in MC pairs remains connected to and misrepresents some aspect of the image, such as depicted entities or events.

Previous studies did not make a distinction between MM and \AMM \ while collecting or annotating their datasets.
Given 200 random samples from COSMOS and following the classification taxonomy of Snopes\footnote{``Fact Checks Rating'' in \url{https://www.snopes.com/sitemap}} we found that 48\% of COSMOS pairs are ``false claims'' (41\% associative imagery and 7\% reinforcing captions) while 52\% were classified as ``miscaptioned'', which we consider to be MM because it implies a relationship between the two modalities. 
After de-duplicating the images of the COSMOS benchmark, the rates were 41\% miscaptioned, 35\% associative imagery, 4\%  reinforcing captions and 20\% duplicates. 
On Fakeddit -given 300 random samples- roughly 45\% of pairs were considered \AMM, with 41\% being manipulated images and 4\% with associative imagery. Moreover, we consider that roughly 14\% of Fakeddit's samples are MM since the remaining 40\% were mostly funny memes, visual jokes, pareidolia imagery and other content that is not generally considered to be misinformation\footnote{The assessment of the COSMOS benchmark follows the taxonomy of Snopes, which is based on the judgement of professional fact-checkers, while the assessment of Fakeddit was conducted by the authors and thus should be interpreted as a rough estimate and not definitive.}.

\begin{figure*}[!t]
\centering
\includegraphics[width=6.5in]{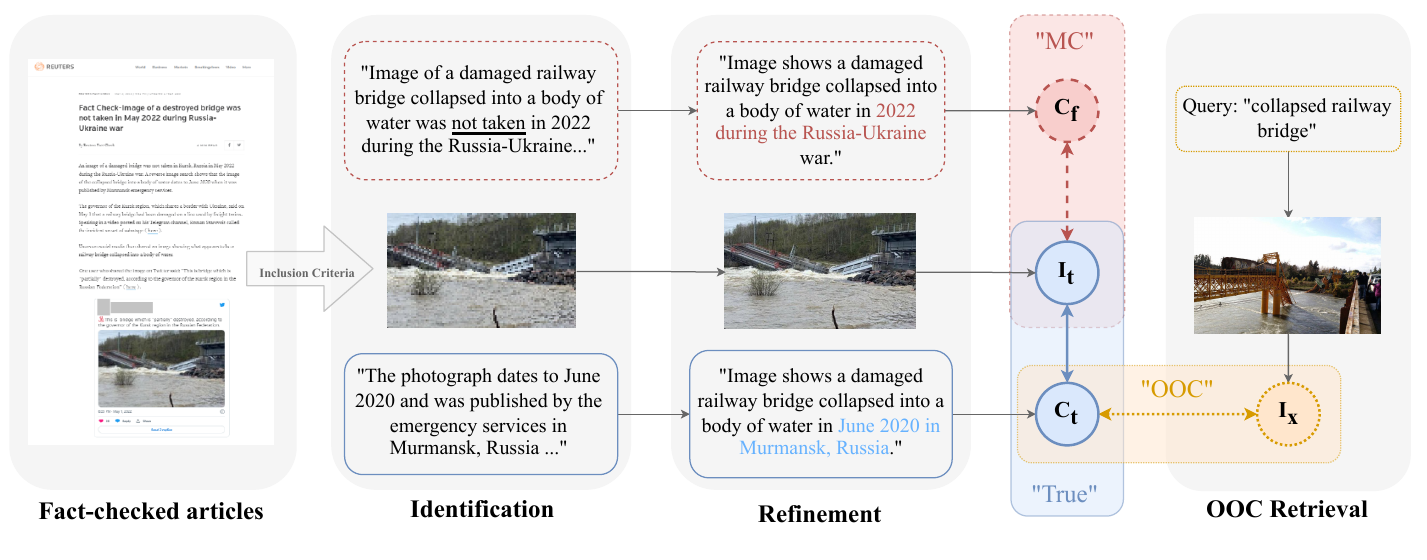}
\caption{Data collection, filtering and refinement process for creating \benchmark.}
\label{fig:verite}
\end{figure*}

\subsection{Creating the \benchmark \ evaluation benchmark}
\label{sec:verite}

Due to the lack of a robust evaluation benchmark for \MMD \ that accounts for unimodal bias, we introduce the ``VERification of Image-TExt pairs'' (\benchmark) benchmark. 
\benchmark \ comprises three classes: True, OOC and MC pairs. 
The data collection process is illustrated in Fig. \ref{fig:verite} and involves the following steps:
\begin{enumerate}
    \item Define inclusion criteria
    \begin{itemize}
        \item  Consider fact-checked articles from 
        Snopes and Reuters that are classified as ``MisCaptioned'' (MC). 
        
        \item Exclude articles classified as ``false claim'', ``legend'', ``satire', ``scam'', ``misattributed'' and other categories that do not adhere to our definition of MM.
        
        \item Exclude articles regarding video footage or animated content and keep image-related cases,
        unless a screenshot of the video is provided that clearly captures the content and claim of the caption. 
        
        \item Include manipulated images (digital art, AI-generated imagery, etc.) only if they are not created with intention to misinform, and their initial origin, content, context, or meaning has been misrepresented within the claim.        
        
    \end{itemize}

    \item Select images and captions
    \begin{itemize}
    
        \item Review the article and collect the misleading claim $C_i^f$.
        
        \item Collect the image $I_i^t$ that is related to $C_i^f$. 

        \item Extract the truthful claim $C_i^t$ for $I_i^t$ from the article.      
        
        \item Examine if claim $C_i^f$ is linked to $I_i^t$ and misrepresents some aspect of it (e.g. origin, content, context, depicted entities etc).If not, exclude for being \AMM.
      
    \end{itemize}
   
    \item Refine captions and images
    \begin{itemize}
        \item Remove ``giveaway'' words such as ``supposedly'', ``allegedly'', ``however'' or phrases like ``this is not the case'', that negate the false claim.      
        Such words and phrases, if learned during the training process, could be used as ``shortcuts'' by \MMD \ models. 
        
        \item Rephrase $C_i^f$ to mimic the syntactic and grammatical structure of $C_i^t$ in order to avoid potential linguistic biases.
        
        \item Rephrase both $C_i^t$ and $C_i^f$ to follow the format: "An image shows.." or "Photograph showing.." in order to create a direct link between the two modalities.
        
        \item Examine both $C_i^t$ and $C_i^f$ for spelling and grammatical errors using ``Google Docs spelling and grammar check''.

        \item Verify that the images are of reasonable quality and do not have any watermarks. If needed, use reverse image search to find the exact same image in better quality.   
        
    \end{itemize}

   \item OOC Image retrieval
    \begin{itemize}

        \item Extract relevant keywords, or their synonyms, in $C_i^t$ to create a query $Q$. 
        
        \item Use Google image search to retrieve one OOC image $I_i^x$ based on $Q$.
        
        \item Ensure that $C_i^t$ and $I_i^x$ share a discernible and meaningful connection (identical or similar origin, content, context or depicted entities) and their alignment is deceptive.
        
    \end{itemize}

\end{enumerate}
To illustrate the aforementioned process in practice, let us consider the example shown in Fig. \ref{fig:verite}.
Starting with a fact-checked article\footnote{\url{https://www.reuters.com/article/factcheck-destroyed-bridge-idUSL2N2WU1CM}}, we collect $I_i^t$ showing a damaged railway that has collapsed into a body of water and $C_i^f$ falsely claiming that the event occurred in ``2022 during the Russia-Ukraine war''. 
We also collect the truthful $C_i^t$ which is provided by professional fact-checkers. $C_i^t$ clarifies that the event took place in ``June 2020 in Murmanask, Russian'' and thus is unrelated to the 2022 Russia-Ukraine war. 
Afterwards, we extract keywords from $C_i^t$ and use $Q=$``collapsed railway bridge'' as the query and retrieve $I_i^x$ from Google Images. 
Similar to $I_i^t$, $I_i^x$ also depicts a collapsed railway bridge but it was captured in Chile, not Russia; thus misaligning the ``location'' entity.

We collected 260 articles from Snopes and 78 from Reuters that met our criteria, which translates to
338 $(I_i^t, C_i^t)$, 338 $(I_i^t, C_i^f)$ and 324 $(I_i^x, C_i^t)$ pairs for True, MC and OOC respectively. 
The collected Snopes articles date as far back as January 2001 up to January 2023, whereas Reuters -only allowing searches up to two years in the past- date from January 2021 to January 2023.
The collected data cover a wide and diverse array of topics and cases including world news (29.04\%), politics (27.94\%), culture and arts (8.82\%), entertainment (7.72\%), sports (3.67\%), the environment (3.66\%), religion (2.94\%), travel (2.57\%), business (2.20\%), science and technology (2.19\%), health and wellness (1.46\%) and others\footnote{To extract and estimate the frequency distribution of news categories, we used \url{https://huggingface.co/Yueh-Huan/news-category-classification-distilbert}}. 

We introduce the term ``modality balancing'' to denote that $I_i^t$ and $C_i^t$ are included twice in the dataset: once with the truthful label and once within the misleading label, as seen in Fig. \ref{fig:verite}. 
More specifically, each image is present once in its truthful pair and once in the MC pair while
each caption is present once in its truthful pair and once in the OOC pair. 
This approach ensures that the model will have to focus on both modalities to consistently discern between factual and misleading I-C pairs. 

\begin{figure*}[!t]
\centering
\includegraphics[width=5.7in]{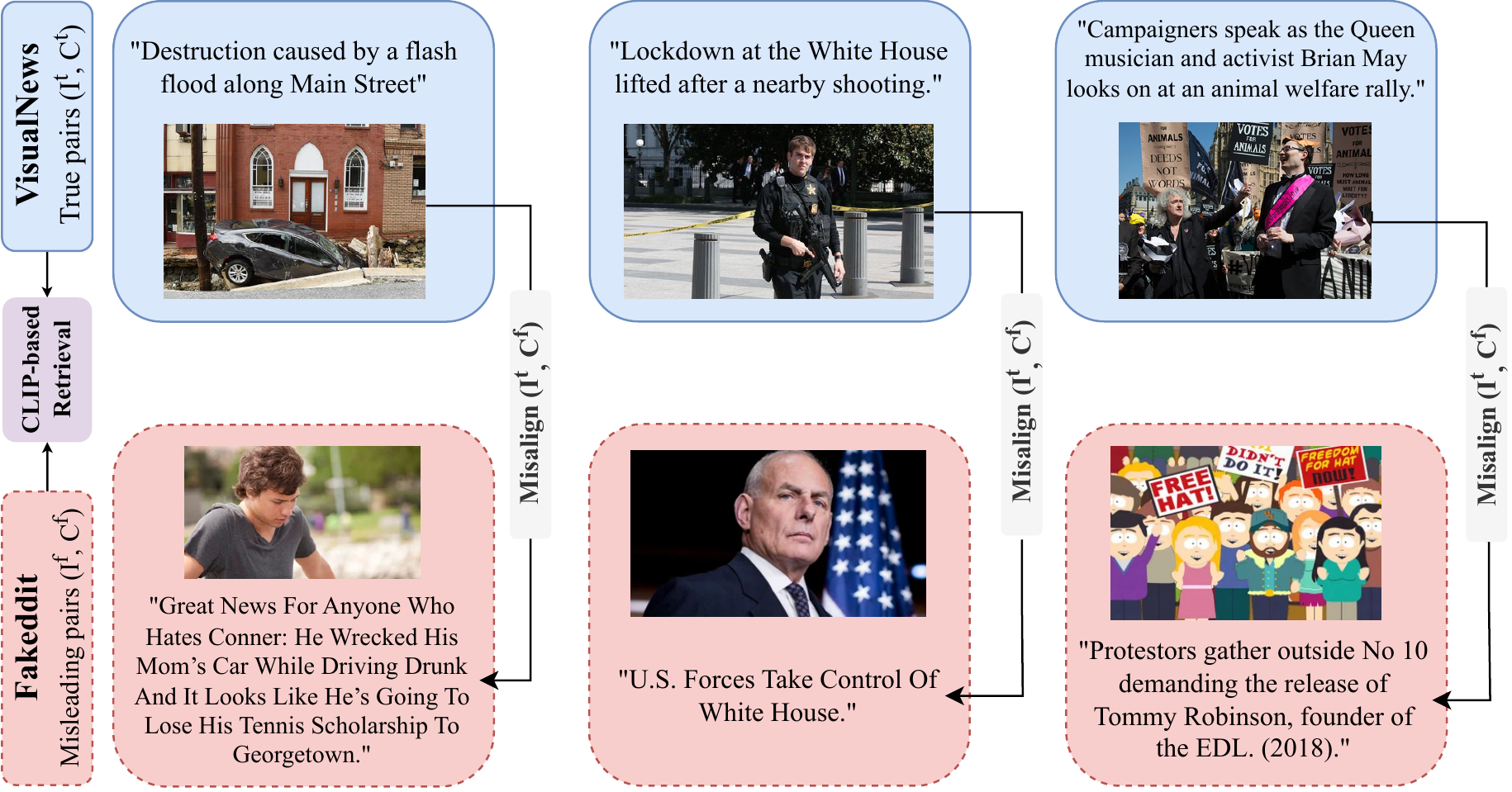}
\caption{Training samples from \methodname \ when applied across the VisualNews and Fakeddit datasets.}
\label{fig:misalign}
\end{figure*}

\subsection{Crossmodal Hard Synthetic Misalignment}
\label{sec:misalign}

Previous studies on synthetic training data for \MMD \ have primarily relied on OOC pairs or NEI.  
These methods create formulaic manipulations, either by re-sampling existing pairs or substituting named entities, and therefore lack the imaginative or expressive characteristic of human produced misinformation such as emotions or irony. Conversely, large weakly annotated datasets may contain noisy labels and high rates of \AMM. 

To address these issues, we propose a new method for generating MM termed Crossmodal HArd Synthetic MisAlignment (\methodname). 
Given a truthful ($I_i^t, C_i^t$) pair and their $V_{I_i^t}$, $T_{C_i^t}$ visual and textual embeddings extracted from CLIP, 
retrieve the most plausible misleading caption $C_j^f$ from a collection of misleading captions 
$\mathcal{C_F}$ with $T_{\mathcal{C_F}}$
textual embeddings, in order to produce a miscaptioned ($I_i^t, C_j^f$) pair with:
\begin{equation}
\label{eqn}
\operatorname*{argmax}\limits_{C_j^f\in \mathcal{C_F}}
\begin{cases}
sim(T_{C_i^t}, T_{C_j^f}),& p\leq 0.5 \\
sim(V_{I_i^t},T_{C_j^f}), & p > 0.5
\end{cases}
\end{equation}
where $p\in[0,1]$ is a uniformly sampled number that determines calculating the cosine similarity ($sim$) between text-to-text or image-to-text pairs.

We apply crossmodal hard synthetic misalignment between VisualNews \cite{liu2020visual} dataset - consisting of  1,259,732 $(I_i^t, C_i^t)$ pairs - and the Fakeddit dataset $(I_j^f, C_j^f)$  \cite{nakamura2019r}.
Out of the 400K misleading captions in $\mathcal{C_F}$ in the Fakeddit dataset, the misalignment process only retains 145,891. 
The resulting generated dataset, termed \textit{\methodname}, is balanced between 1.2M $(I_i^t, C_i^t)$ truthful and 1.2M $(I_i^t, C_j^f)$ miscaptioned pairs. 
Since $C_j^f$ from Fakeddit may have been aligned with more than one image from VisualNews, we also create \textit{\methodname-D} by removing duplicate instances of $C_j^f$. 
We balance the classes of \textit{\methodname-D} through random down-sampling. 
The resulting dataset consists of 145,891 $(I_i^t, C_i^t)$ and an equal number of $(I_i^t, C_j^f)$.

We randomly sample 100 instances from the generated data and determined that approximately 73\% of generated $(I_i^t, C_j^f)$ can be considered MM while 12\% are \AMM.
Moreover, 6\% of the pairs in the dataset are accidentally correct pairs, for instance, an image of firefighters near a fire being paired with the caption ``Firemen battling a blaze''. 
Finally, 9\% of pairs are unclear, containing click-bait captions as 
``You'll never guess how far new home prices have dropped'' which are  paired with a weakly relevant image and cannot necessarily be considered misinformation.
Naturally, the proposed method is not perfect, with approximately 27\% of its samples not aligning with our definition of MM. 
Nevertheless, it provides a significant improvement over the original Fakeddit dataset where roughly 45\% are \AMM \ and only 14\% are MM. 

As seen in the examples of Fig. \ref{fig:misalign} (bottom), misleading captions $C_j^f$ from Fakeddit can contain humor, irony and be more imaginative than named entity substitutions. 
However, their connections with the images $I_j^f$ are often \AMM \ or can be easy to detect (e.g. an illustrated image being humorously paired with a real demonstration).
Conversely, \textit{\methodname} maintains the `desired' aspects of $C_j^f$ (e.g. sarcasm, emotions, etc.) but pairs them with more relevant imagery, thus creating ``hard'' samples and by extension more robust training data.
For example, consider the case shown in Fig. \ref{fig:misalign}, where an illustrated image is humorously paired with a caption about a demonstration and is subsequently `misaligned' with an image of a real protest, thus creating a more  realistic misleading pair.

In contrast to NEI-based methods, our generated samples consists of human-written misinformation rather than simple named entity manipulations.
Finally, unlike NewsCLIPings, \methodname \ utilizes CLIP-based retrieval to generate MC rather than OOC pairs and employs both intra-modal and cross-modal similarity to create synthetic samples. 

\subsection{Detection model}

In our experiments, we encode all image-caption pairs $(I,C)$ using the pretrained CLIP ViT-L/14 \cite{radford2021learning} both as the image encoder $E_{I}(\cdot)$ and the textual encoder $E_{C}(\cdot)$ that produce the corresponding vector representations 
$V_{I}\in{R}^{m\times 1}$ and $T_{C}\in{R}^{m\times 1}$ respectively, where $m=768$ the encoder's embedding dimension. 
CLIP is an open and widely used model for multimodal feature extraction in numerous multimedia analysis and retrieval tasks \cite{lin2022frozen, guzhov2022audioclip, li2022clip} including multimedia verification and has yielded promising results \cite{papadopoulos2023synthetic, luo2021newsclippings,zhang2023ino, zhou2023multimodal, tahmasebi2023improving}. 

We concatenate the extracted features across the first or `token' axis as $[ V_{I}, T_{C}] \in{R}^{m\times 2} $. 
(`batch dimension' omitted for clarity).
As the ``detector'' $D(\cdot)$ we use the Transformer encoder \cite{vaswani2017attention} 
but exclude positional encoding and use average pooling instead of a CLS token.
$D(\cdot)$ comprises $L$ layers of $h$ attention heads and a feed-forward network of $f$ dimension and outputs $y$:
\begin{equation}
y = \textbf{W}1\cdot \text{GELU}(\textbf{W}0\cdot \text{LN}(\text{D}([V_{I}, T_{C}])))
\end{equation}
where LN stands for Layer Normalization, $\textbf{W}_0\in\mathbb{R}^{m \times 2}$ 
is a GELU activated fully connected layer and 
$\textbf{W}_1\in\mathbb{R}^{n \times \frac{m}{2}}$ 
is the final classification layer with $n=1$ for binary and $n=3$ for multiclass tasks (learnable bias terms are considered but omitted here for clarity). The network is optimized based on the categorical cross entropy or the binary cross entropy loss function for multiclass or binary tasks, respectively. 

For unimodal experiments we only pass $V_{I}$ or $T_{C}$ through $D(\cdot)$ and define $\textbf{W}_0\in\mathbb{R}^{m \times 1}$. 
In these cases, $D(\cdot)$ only receives a single input token. 
Therefore, its attention scores are uniformly assigned a value of 1, resulting in an absence of distinct attention weights.
We denote this ``Transformer'' detector as $D\textsuperscript{-}(\cdot)$; $D(\cdot)$ minus multi-head self-attention, 
since the latter has no contributing role.

Moreover, in order to investigate the role that multi-head self-attention
plays in unimodal bias, we conduct additional experiments using the variant $D\textsuperscript{-}(\cdot)$ where the two modalities are concatenated along the second or ``dimensional'' axis, resulting $[ V_{I} ; T_{C}]\in\mathbb{R}^{2m\times 1}$.

\section{Experimental Setup}

\subsection{Training datasets and competing methods}

First, we train $D(\cdot)$ on the \textbf{\twitterdataset} (MediaEval 2016\footnote{\url{https://github.com/MKLab-ITI/image-verification-corpus}}) dataset and compare it against numerous \MMD \ models, namely: Event Adversarial Neural Network (\textit{EANN}) using VGG-19 and Text-CNN \cite{wang2018eann}, Multimodal Variational Autoencoder (\textit{MVAE}) using VGG-19 and Bi-LSTMs \cite{khattar2019mvae}, 
\textit{SpotFake} using VGG-19 and BERT \cite{singhal2019spotfake}, Bidirectional Crossmodal Fusion (\textit{BCMF}) network using DeiT and BERT \cite{yu2022bcmf} and a transformer-based architecture employing Faster-RCNN and BERT to capture intra-modal relations and a multiplicative multimodal method to capture inter-modal relations (\textit{Intra+Inter}) \cite{singhal2022leveraging}.

Afterwards, we compare $D(\cdot)$ when trained on the original \textbf{Fakeddit} \cite{nakamura2019r}, our \textbf{\methodname} and \textbf{\methodname-D} datasets as well as numerous synthetically generated datasets, including OOC: 
NewsCLIPings text-text (\textbf{NC-t2t}) \cite{luo2021newsclippings}, 
random-sampling by topic (\textbf{RSt}) \cite{papadopoulos2023synthetic} as well as NEI: 
MEIR \cite{sabir2018deep}, 
random named entity swapping by topic (\textbf{R-NESt}) and CLIP-based named entity swapping by topic (\textbf{CLIP-NESt}) \cite{papadopoulos2023synthetic}. 
The number of samples per class for each dataset can be seen in Table \ref{table:data_stats}.

Furthermore, we experiment with dataset aggregation, the combination of various generated datasets.
Aggregated datasets are denoted with a plus sign, for instance \textit{R-NESt + NC-t2t}.
For the multiclass task, we combine one OOC dataset and at least one MC dataset to represent the OOC and MC classes, respectively.
To evaluate the contribution of \methodname \ (or \methodname-D) in \MMD \, we perform an ablation experiment where they are either integrated or excluded from aggregated datasets.
Note that, during training, we apply random down-sampling to address any class imbalance.

\begin{table}
\centering
\caption{
Number of samples per class in each training and testing dataset. ``*'' denotes datasets whose ``false'' pairs exhibit more similarities to, but may not entirely align with, our definition of miscaptioned (MC) images. Validation sets are used but omitted here.
}
\label{table:data_stats}
  \begin{tabular}{llll}

    \toprule 
    \textbf{Training Dataset} &
    \textbf{True} &
    \textbf{OOC} &
    \textbf{MC}
    \\
    
    \midrule

     VisualNews & 1,007,744 & - & - \\

    
     RSt & 
     1,007,744 & 1,007,744 & - \\
     
     NC-t2t & 
     258,036 & 258,036 & - \\

     CSt &
     1,007,744 & 1,007,744 & - \\ 

     
     MEIR &
     82,156 & - & 57,940 \\
     
     CLIP-NESt & 
     1,007,744 & - &  847,693 \\
     
     R-NESt & 
     1,007,744 & - & 924,586 \\     

     
     Fakeddit*
     & 267,601 & - &  413,197 \\
     
     \methodname \ & 
     1,007,744 & - & 1,007,744 \\
     
     \methodname-D & 
     145,891 & - & 145,891 \\

     \twitterdataset* & 7,292 & - & 9,148 \\    

     \midrule

    \textbf{Testing Dataset} & & & \\     
    \midrule
     \twitterdataset* & 467 & - & 623 \\
     COSMOS* & 850 & - & 850 \\
     \benchmark & 338 & 324 & 338 \\
     
    \bottomrule
\end{tabular}
\end{table}

\begin{figure}[!t]
\centering
\includegraphics[width=4in]{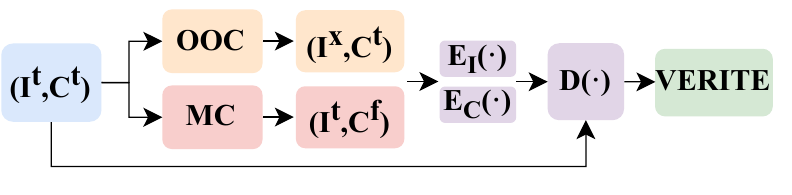}
\caption{
High-level overview of the employed pipeline. 
}
\label{fig:pipeline}
\end{figure}

Figure \ref{fig:pipeline} presents a high-level overview of our employed pipeline. We incorporate truthful image-caption pairs from the VisualNews dataset and employ an OOC-based (e.g. NewsCLIPings) and a MC-based generation method (e.g. \methodname) to create false OOC and MC pairs, respectively. Subsequently, we utilize CLIP to extract the visual and textual features from the image-caption pairs and then train the multiclass Transformer detector $D(\cdot)$ before ultimately assessing its performance on the \benchmark \ benchmark.

\subsection{Evaluation protocol}
\label{sec:eval_protocol}

Considering the distribution shift between training (generated) and test sets (real-world), utilizing an ``out-of-distribution'' validation set could potentially result in slightly better test accuracy \cite{koh2021wilds}. 
However, due to the relatively small sizes of both COSMOS and \benchmark \ datasets, we decided to avoid this approach.
Instead, after training, we retrieve the best performing hyper-parameter combination based on the ``in-distribution'' validation set (generated) and evaluate it on the final test (real-world) sets: COSMOS and \benchmark.
For evaluation, we report the accuracy score (image-only, text-only or multimodal) for binary classification on COSMOS and multiclass accuracy on \benchmark.
Moreover, we experiment with a binary version of \benchmark \ (\benchmark-B) where both ``OOC'' and ``MC'' pairs are combined into a single class denoting misinformation.  
Here, we report the accuracy for each pair of classes, namely ``True vs OOC'' and ``True vs MC''.
The number of samples per class for each evaluation dataset can be seen in Table \ref{table:data_stats}.

Prior works using the \twitterdataset \ dataset do not specify the validation set used for hyperparameter tuning \cite{khattar2019mvae, singhal2019spotfake, yu2022bcmf}. 
By inspecting their code\footnote{\url{https://github.com/dhruvkhattar/MVAE}} \footnote{\url{https://github.com/shiivangii/SpotFake}},
we can deduce that the test set was used for this purpose, which is problematic.  
We follow this protocol only for comparability and also train $D(\cdot)$ using a corrected protocol, where the development set is randomly split into training (90\%) and validation (10\%). 

To evaluate the presence and magnitude of unimodal bias, we employ two metrics: the percentage increase in accuracy ($\Delta\%$) between a unimodal model and its multimodal counterpart, and Cohen's d ($d$) effect size.
Negative $\Delta\%$ and positive $d$ values serve as indicators for the presence of unimodal bias.

\subsection{Implementation details}
$D(\cdot)$ is trained for a maximum of 30 epochs (early stopping at 10 epochs) by the Adam optimizer with a learning rate of $lr = 5e-5$. 
For tuning the hyperparameters of $D(\cdot)$ consider the following values: $L \in \{1, 4\}$ transformer layers of $f \in \{128, 1024\}$ dimension of the feed-forward network model, $h \in \{2, 8\}$ attention heads. 
The dropout rate is constant at 0.1 and the batch size at 512. 
This grid-search results in a total of 8 experiments per modality (image-only, text-only, multimodal), thus 24 per dataset.
For experiments on the \twitterdataset \ dataset, we reduce the batch size to 16 and define $lr \in \{5e-5, 1e-5\}$, since it is a much smaller dataset.
We set a constant random seed (0) for Torch, Python random and NumPy to ensure the reproducibility of our experiments. 
We conducted the experiments on a computer equipped with an AMD Ryzen 3960X 24-Core CPU, 128GB of RAM, and a single GeForce RTX 3060 GPU.

\section{Experimental Results}

\textit{\textbf{Image-side unimodal bias on \twitterdataset :}}

We begin by comparing the performance of $D(\cdot)$ with various models trained and evaluated on the \twitterdataset \ dataset. 
In Table \ref{table:twitter}, we observe that among multimodal models, $ D\textsuperscript{-}(I;C) $ achieves the third-highest result (80.5\%), 
after Intra+Inter (83.1\%) and BCMF (81.5\%).
However, it is noteworthy that the image-only model $ D\textsuperscript{-}(I) $ achieves the highest overall accuracy (83.7\%). This finding indicates the presence of image-side unimodal bias within models trained and evaluated on the \twitterdataset.
Table \ref{table:unimodal_bias} also demonstrates that $ D(I,C) $ displays a greater percentage decrease (-4.78\%) compared to $ D\textsuperscript{-}(I;C) $ (-3.92\%), thus \twitterdataset \ does not seem to allow the full utilization of multi-head self-attention. 

\begin{table}
\centering
\caption{
Performance of Transformer $D(I,C)$ and $D\textsuperscript{-}(\cdot)$ 
for caption-only $(C)$, image-only $(I)$ or multimodal inputs $(I;C)$ when trained and evaluated on the \twitterdataset \ dataset.
\textbf{Bold} denotes the highest binary accuracy.
}
\label{table:twitter}
  \begin{tabular}{lllc}

   \toprule
   
    \textbf{Model} 
    & \textbf{$E_I(\cdot)$} 
    & \textbf{$E_C(\cdot)$}
    & \textbf{Accuracy}  \\
    
    \midrule 

    EANN \cite{wang2018eann} & VGG-19 & TextCNN & 71.5 \\
    MVAE \cite{khattar2019mvae} & VGG-19 & BiLSTM & 74.5 \\
    SpotFake \cite{singhal2019spotfake} & VGG-19 & BERT & 77.8 \\
    BCMF \cite{yu2022bcmf} & DeiT & BERT & 81.5 \\
    Intra+Inter \cite{singhal2022leveraging} & Faster-RCNN & BERT & 83.1 \\

    \midrule

    $D\textsuperscript{-}(C)$ & - & CLIP & 74.7 \\
    $D\textsuperscript{-}(I)$ & CLIP & - & \textbf{83.7} \\
    $D\textsuperscript{-}(I;C)$ & CLIP & CLIP & 80.5 \\ 
    $D(I,C)$ & CLIP & CLIP & 79.7 \\
    
\bottomrule 
\end{tabular}
\end{table}

\begin{figure}[!t]
\centering
\includegraphics[width=4in]{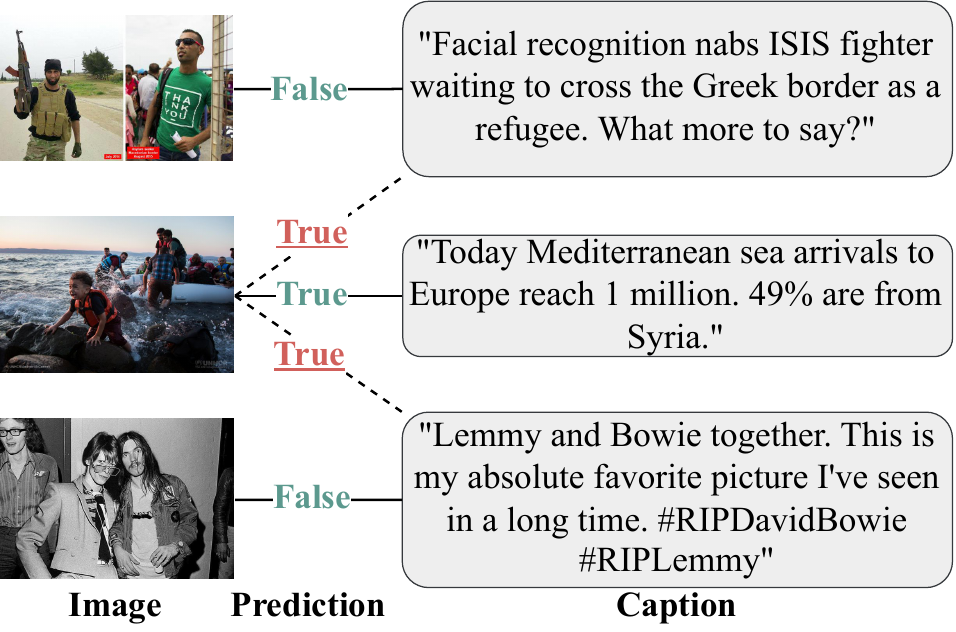}
\caption{Inference by $D(I,C)$ on three samples from \twitterdataset. Moreover, we examine the model's image-side unimodal bias by inputting the middle image along with each of the three captions. $D(I,C)$ predicts ``true'' with all three captions, which means that the model does not take the caption into consideration. 
Red underlines denote mistaken predictions. 
}
\label{fig:inference_twitter}
\end{figure}

Fig. \ref{fig:inference_twitter} demonstrates that the multimodal model $D(I,C)$ produces the same outputs regardless of whether the image is paired with its corresponding caption or two randomly selected captions. 
$D(I,C)$ predicts that all pairs are ``true'' regardless of the accompanying text.
This example visually highlights the presence of image-side unimodal bias within the model's inference process.

The occurrence of image-side unimodal bias can be attributed to two primary factors. Firstly, \twitterdataset \ was originally designed as an image verification corpus, comprising a substantial number of manipulated or edited images. Consequently, the significance of the accompanying text diminishes, as the primary source of misinformation lies within the image itself, what we term \AMM.
Secondly, \twitterdataset \ exhibits an imbalance between the number of texts and images used for training and testing. With only 410 images available for training and 104 images for testing, compared to approximately 17k and 1k tweets respectively, each image appears multiple times in the dataset, albeit with different texts. This discrepancy can lead to the model disregarding the textual modality, further reinforcing the image-side bias.
Considering these factors, it appears that \twitterdataset \ may not be an optimal choice for training and evaluating models for the task of \MMD \ and might be better suited for its original purpose, namely, image verification.

As discussed in Section \ref{sec:eval_protocol}, it is worth highlighting that the evaluation protocol employed in \cite{khattar2019mvae, singhal2019spotfake, yu2022bcmf} is problematic; using the test set during the validation and/or hyper-parameter tuning process.
Under the corrected evaluation protocol, $D\textsuperscript{-}(I)$ achieves 81.0\% accuracy, $D\textsuperscript{-}(I;C)$ achieves 77.3\% (-4.56\%), and $D(I,C)$ achieves 76.66\% (-5.35\%). 
The aforementioned conclusions regarding image-side bias remain consistent even under the corrected evaluation protocol.

Finally, note that a direct comparison between the models in Table \ref{table:twitter} is not possible as they employ different image and text encoders. Consequently, we refrain from asserting that we have attained ``state-of-the-art'' performance on the \twitterdataset. 
Instead, the results showcase that $D(\cdot)$ can provide competitive and reasonably strong performance -while being a relatively simple architecture- and will be leveraged in all preceding  experiments.

\textit{\textbf{Text-side unimodal bias on COSMOS:}}
We proceed by training $ D(\cdot) $ on various datasets for binary classification and evaluating on the COSMOS benchmark, as illustrated in Table \ref{table:cosmos}.
We observe that the text-only $ D\textsuperscript{-}(C) $ trained on \textit{\methodname-D} achieves 72.6\% accuracy on COSMOS, the highest accuracy score on COSMOS. 
However, this translates into the text-only model outperforming its multimodal counterparts, $ D\textsuperscript{-}(I;C) $ and $ D(I,C)$ by -7.85\% and -14.88\% respectively. 
As seen in Table \ref{table:unimodal_bias}, on average, $ D\textsuperscript{-}(C) $ outperforms $ D\textsuperscript{-}(I;C) $ by 2.34\% and $ D(I,C)$ by 3.47\% with a $d$ of 0.25 and 0.4 respectively; highlighting that COSMOS does not seem to allow the full utilization of multi-head self-attention.
We also observe in Table \ref{table:cosmos} that both $ D(I,C) $ and $ D\textsuperscript{-}(I;C) $ suffer from text-side unimodal bias on COSMOS, only when trained with NEI-based datasets (CLIP-NESt and R-NESt) or datasets relying on human-written misinformation (Fakeddit and \methodname). 
Text-manipulations and human-written texts may display certain  linguistic patterns that inadvertently the models learn to attend to while reducing attention towards the visual modality.

\begin{figure}
\centering
\includegraphics[width=4in]{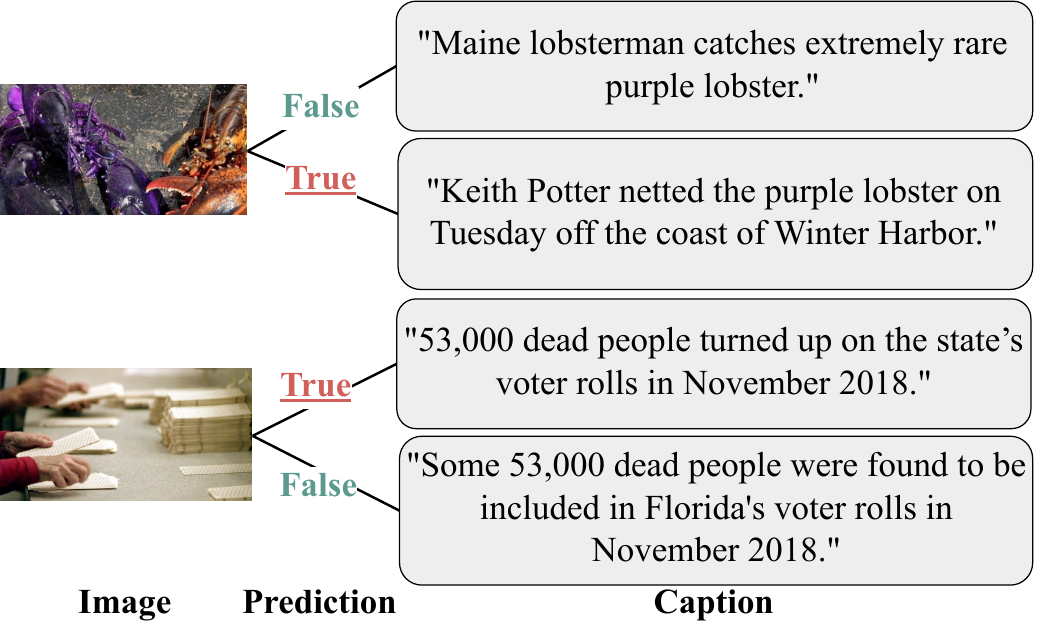}
\caption{Inference on two misleading samples from COSMOS with near-duplicate texts by $D(I,C)$ trained on \methodname-D.
Red underlines denote mistaken predictions. 
}
\label{fig:inference_cosmos}
\end{figure}

Fig. \ref{fig:inference_cosmos} provides a visual representation of the behavior of the multimodal model $D(I,C)$ when trained on \methodname-D and evaluated on COSMOS. 
It showcases that the model can generate different outputs when applied to near-duplicate image-caption pairs, where the textual content exhibits only very minor differences that do not significantly alter the fact that it represents misinformation.
Considering these results, we can conclude that COSMOS is not an ideal choice when it comes evaluating models for the task of multimodal misinformation detection.
The dataset's characteristics and composition allow for the presence and reinforcement of text-sided unimodal bias, thereby yielding misleading or falsely optimistic outcomes.

\begin{table}
\centering
\caption{Results on the COSMOS benchmark. We report the performance of Transformer $D(I,C)$ and $D\textsuperscript{-}(\cdot)$ for
caption-only (C), image-only (I) or multimodal inputs
(C; I). 
\textbf{Bold} denotes the highest binary accuracy.
}
\label{table:cosmos}
  \begin{tabular}{lcccc}

    \toprule 
    \textbf{Training} &
    \textbf{$D\textsuperscript{-}(I)$} &    
    \textbf{$D\textsuperscript{-}(C)$} & \textbf{$D\textsuperscript{-}(I;C)$} &
    \textbf{$D(I,C)$}
    \\
    
    \midrule

     RSt & 
     50.0 & 
     50.0 & 
     51.1 &
     51.5
     \\
     
     NC-t2t & 
     45.1 &
     50.0 & 
     50.8 &
     52.6 
     \\

     CSt &
     45.5 &
     47.4 & 
     52.6 &
     \\   

     \midrule
     
     MEIR &
     50.7 &
     51.9 & 
     52.6 &
     53.2
     \\
     
     CLIP-NESt & 
     50.0 &
     55.4 & 
     53.6 &
     53.4
     \\
     
     R-NESt & 
     50.0 &
     59.5 & 
     54.1 &
     55.2	
     \\     

     \midrule
     
     Fakeddit 
     & 57.4	
     & 61.7 
     & 57.9 
     & 52.5
     \\
     
     \methodname \ & 
     64.5 &
     67.6 & 
     60.3 &
     58.9
     \\
     
     \methodname-D & 
     64.4 &
     \textbf{72.6}
     &  
     66.9 &
     61.8
     \\
    
\bottomrule
\end{tabular}
\end{table}

\textit{\textbf{Unimodal bias is not (entirely) algorithmic:}}
Table \ref{table:testset} presents the performance of $ D(\cdot) $ when trained on various datasets and evaluated on their respective test sets. When trained on OOC-based datasets (RSt, NC-t2t, and CSt) $ D(\cdot) $ performs poorly in both image- and text-only settings -with an average of 53.6\% and 52.5\% respectively- while achieving high multimodal accuracy. 
Expectedly, as both the image and the caption in OOC samples are factually accurate, but only their relation is corrupted, it is not possible to determine the existence of misinformation by solely examining one modality.

In contrast, $ D\textsuperscript{-}(C) $ trained on NEI datasets (MEIR, R-NESt, CLIP-NESt) and \methodname \ perform in closer proximity to the multimodal one, with $D\textsuperscript{-}(C)$ scoring 81.4\%, compared to 86.6\% by $D\textsuperscript{-}(I;C)$ and 85.9\% by $D(I,C)$.
At the same time, the image-only setting yields significantly lower performance for NEI methods and \textit{\methodname}; the only exception being \textit{Fakeddit}, which comprises a higher percentage of manipulated images.
Once again, these results suggest that methods relying on text manipulation or human-written misinformation may introduce linguistic patterns and biases that render the image less important.

However, unlike the COSMOS benchmark, no unimodal method surpasses its multimodal counterparts on the test sets. 
This is also demonstrated in Table \ref{table:unimodal_bias}, where neither $\Delta\%$ nor $d$ indicate the presence of any unimodal bias. 
We can deduce that unimodal bias is partially algorithmic -an \MMD \ model may rely on certain superficial unimodal patterns during training- 
but more importantly, these biases are significantly exacerbated by certain characteristics of \twitterdataset and COSMOS -one of which is the high prevalence of \AMM \ instances- thus raising concerns about their reliability as evaluation benchmarks.

\begin{table}
\centering
\caption{Binary classification results on the test set of each dataset.
}
\label{table:testset}
  \begin{tabular}{lcccc}

    \toprule 
    \textbf{Training Dataset} & \textbf{$D\textsuperscript{-}(C)$} & \textbf{$D\textsuperscript{-}(I)$} & \textbf{$D\textsuperscript{-}(I;C)$} & 
    \textbf{$D(I,C)$}     
    \\
    
    \midrule

     RSt & 50.0 & 50.0 & 96.5 & 96.6 \\
     CSt & 57.4 & 57.2 & 75.1 & 75.9	\\
     NC-t2t & 50.0 & 53.7 & 84.0 & 84.3 \\
     
     \midrule
     MEIR & 72.2 & 65.5 & 76.1 & 73.9 \\
     R-NESt & 82.6 & 52.1 & 91.0 & 91.2\\
     CLIP-NESt & 65.5 & 54.0 & 70.4 & 70.3 \\
     
     \midrule
     Fakeddit & 90.9 & 90.1 & 95.1 & 94.5 \\
     \methodname & 90.1 & 50.0 & 93.0 & 91.3 \\
     \methodname-D & 86.9 & 60.4 & 94.1 & 94.3 \\
    
\bottomrule 

\end{tabular}
\end{table}

\textbf{\benchmark \ alleviates unimodal bias:}
The analysis of Table \ref{table:unimodal_bias} reveals that both $\Delta\%$ and Cohen's $d$ effect sizes indicate the absence of any unimodal bias on the \benchmark \ benchmark. Notably, $ D(I,C) $ displays an average 27.94\% increase in accuracy when compared to text-only $ D\textsuperscript{-}(C) $ and 43.27\% when compared to image-only $ D\textsuperscript{-}(I) $. These results emphasize that a model biased towards one modality can not achieve satisfactory performance on \benchmark.
Furthermore, it is worth noting that $ D(I,C) $ consistently outperforms $ D\textsuperscript{-}(I;C) $, demonstrating that \benchmark \ effectively allows for the power of 
multi-head self-attention
to be leveraged, unlike COSMOS and \twitterdataset. 

Additionally, we train $D(\cdot)$ for binary classification and evaluate its performance on \textit{\benchmark-B}; the binary version of \textit{\benchmark}. 
The primary aim of these experiments is to investigate the implications of removing ``modality balancing'' from \textit{\benchmark} in relation to unimodal bias.
This entails that each image no longer appears twice in \textit{\benchmark}, once in the ``True'' class and once in the ``Miscaptioned'' class, and each caption no longer appears twice, once in the ``True'' class and once in the ``out-of-context'' class; since they are separated into two separate evaluations.
In Table \ref{table:verite_binary} we observe that $ D\textsuperscript{-}(I;C) $ trained on R-NESt or \methodname-D exhibits minor instances of unimodal bias in the ``True vs MC'' evaluation. 
The scale of this bias becomes more pronounced when 
multi-head self-attention
is employed in $D(I,C)$. Additionally, when trained with \textit{Fakeddit}, $D(I,C)$ showcases unimodal bias within the ``True vs OOC'' metric.
These findings bear similarities to the patterns identified within the COSMOS benchmark, albeit at a smaller scale, presumably due to the lack of \AMM \ in \benchmark.
Based on these results, we can infer that ``modality balancing'' plays a crucial role in mitigating the manifestation of unimodal bias within \textit{\benchmark}. 
Hence, we advise against employing \textit{\benchmark-B} as an evaluation benchmark for multimodal misinformation detection; especially of MC pairs.
Instead, we recommend utilizing the original \textit{\benchmark} benchmark, as it has demonstrated its robustness as a comprehensive evaluation framework. 

\begin{table*}
\centering
\caption{Multiclass classification results on the \textit{\benchmark} dataset with different training MC data. For OOC data, NC-t2t is used in all experiments. }
\label{table:verite}
  \begin{tabular}{lccccccc}

    \toprule 
    &
    \multicolumn{4}{c}{\textbf{\boldmath$ D(I,C)$}}
    & \boldmath$ D\textsuperscript{-}(I;C)$
    & \boldmath$ D\textsuperscript{-}(C)$
    & \boldmath$ D\textsuperscript{-}(I)$ \\
    
    \textbf{MC Data} & 
    \textbf{Accuracy} & \textbf{True} & \textbf{MC} & \textbf{OOC} & \textbf{Accuracy} & \textbf{Accuracy} & \textbf{Accuracy} \\

    \midrule

     CLIP-NESt & 
     47.7 & 64.5 &	\textbf{35.2} &	43.2 & 40.1 & 33.1 & 33.8 \\
         
     R-NESt & 
     47.7 & 79.0 &	19.2 & 44.8 & 43.9 & 33.6 & 34.6 \\
     
     \methodname & 
     48.7 & 79.0 & 16.3 & 50.9 & 47.9 & 37.3 & 34.4 \\

     \methodname-D & 
     49.0 & 81.7 &	13.0 & \textbf{52.5} & 47.7 &  40.6 & 37.5 \\     
     
     \midrule
          
     R-NESt+\methodname & 
     49.6 & 76.9 & 23.1 & 48.8 & 51.1 & 39.5 & 34.8 \\

     R-NESt+\methodname-D & 
     50.0	& 80.2 & 24.3 & 45.4 & 46.9 & 41.7 & 33.2 \\

     CLIP-NESt+\methodname & 
     50.8 & \textbf{83.7} & 21.0 &	47.5 & 49.6
     & 41.8 & 33.4
     \\

     CLIP-NESt+\methodname-D & 
     \textbf{52.1} & 70.7 & 33.4 & 52.2 & 49.3 & 43.7 & 34.8 \\
     
\bottomrule 
\end{tabular}
\end{table*}

\begin{table*}
\centering
\caption{Results on \textit{\benchmark-B} by $ D(\cdot) $ trained on different datasets for binary classification. The objective of these experiments is to investigate the impact on unimodal bias when eliminating ``modality balancing'' from \textit{\benchmark}. 
Evaluation metrics used include ``True vs OOC" and ``True vs MC'' accuracy. In parentheses, we report the percentage improvement ($\Delta\%$) of each multimodal model compared to the text-only model. \textbf{Bold} denotes the best performance per evaluation metric.}
\label{table:verite_binary}
  \begin{tabular}{lccc|ccc}

    \toprule 
     
    & \multicolumn{3}{c}{\textbf{True vs OOC}} 
    & \multicolumn{3}{c}{\textbf{True vs MC}} 
     \\
    
    \midrule

    \textbf{Training Dataset} 
    &  $ D\textsuperscript{-}(C) $ & $ D\textsuperscript{-}(I;C) $   & $ D(I,C) $  
    &  $ D\textsuperscript{-}(C) $ & $ D\textsuperscript{-}(I;C) $   & $ D(I,C) $  
    \\

    \midrule

     Fakeddit & 
     50.4 & 51.5 (2.2) & 48.3 (-4.2) &
     58.7 & 55.9 (-4.8) & 53.6 (-8.7)
     \\
     
     \methodname-D & 
     50.4 & 52.6 (4.4) & 52.0 (3.2) &
     \textbf{64.8} & 64.5 (-0.5) & 58.4 (-9.9) \\
          
     R-NESt   
     & 50.0 & 66.2 (32.4) & 67.2 (34.4)
     & 59.2 & 59.6 (0.68) & 58.6 (-1.0)
     \\

    NC-t2t   
     & 46.5 & 72.4 (55.7) & 72.0 (54.8)
     & 50.0 & 54.4 (8.8) & 54.6 (9.2)
     \\
               
     \midrule

    R-NESt + \methodname-D \ + NC-t2t &
    50.6 & 72.4 (43.1) & \textbf{72.7} (42.8) & 
    58.4 & 63.9 (9.4) & 61.2 (1.3)
    \\
    
\bottomrule 
\end{tabular}
\end{table*}

\begin{table*}
\centering
\caption{
Examination of unimodal bias on different evaluation datasets. 
We report the average percentage increase in terms of accuracy ($\Delta\%$) and the average effect size measured by Cohen's d ($d$). 
Negative $\Delta\%$ and positive $d$ values indicate the presence and magnitude of unimodal bias (denoted with \textbf{bold}).
}
\label{table:unimodal_bias}
  \begin{tabular}{lcccc|cccc}

    \toprule 

    \textbf{Multimodal}
    & 
    \multicolumn{4}{c}{
    \boldmath$ D(I,C)$
    }  

    & \multicolumn{4}{c}{
    \boldmath$ D\textsuperscript{-}(I;C)$ 
    }  

    \\


    \textbf{vs Unimodal}
    & \multicolumn{2}{c}{
    \boldmath$ D\textsuperscript{-}(C)$ 
    }  
    & \multicolumn{2}{c}{
    \textbf{vs} \boldmath$ D\textsuperscript{-}(I)$ 
    }
    & \multicolumn{2}{c}{
      
    \boldmath$ D\textsuperscript{-}(C)$ 
    }
    & \multicolumn{2}{c}{
      
    \textbf{vs} 
    \boldmath$ D\textsuperscript{-}(I)$ 
    } 
    \\
    
    \midrule
    
    \textbf{Dataset} 
    & $\Delta\%$ & $d$
    & $\Delta\%$ & $d$
    & $\Delta\%$ & $d$
    & $\Delta\%$ & $d$
    \\

    \midrule
    
        Test sets & 25.33 & -2.33 & 49.39 & -1.02 & 25.67 & -2.36 & 49.92 & -1.05 \\ 
        COSMOS & \textbf{-3.47} & \textbf{0.41} & 4.09 & -0.28 & \textbf{-2.34} & \textbf{0.25} & 5.32 & -0.39 \\ 
        \twitterdataset & 6.69 & - & \textbf{-4.78} & - & 7.76 & - & \textbf{-3.82} & - \\ 
        \benchmark & 27.94 & -3.56 & 43.27 & -10.41 & 21.38 & -2.19 & 36.28 & -4.68 \\ 
    
    \bottomrule 
    \end{tabular}
\end{table*}

\textbf{\textit{On the performance of \methodname:}}
Table \ref{table:verite} provides a detailed overview of the results obtained on the \benchmark \ evaluation benchmark.
In our training process for multiclass misinformation detection, we employ $D(\cdot)$ using one out-of-context (OOC) dataset in combination with at least one NEI-based or \methodname \ dataset, or both.
We observe that $D(I,C)$ trained on \textit{\methodname \ + NC-t2t}(48.7\%) or \textit{\methodname-D + NC-t2t}(49\%) outperform both \textit{CLIP-NESt + NC-t2t}(47.7\%) and \textit{R-NESt + NC-t2t}(47.7\%).
Furthermore, when $D(I,C)$ is trained on aggregated datasets that include \textit{\methodname}. it consistently outperforms those that do not. Notably, \textit{CLIP-NESt + Misalign + NC-t2t} achieves the highest overall multiclass accuracy of 52.1\%, representing a 9.22\% improvement over \textit{CLIP-NESt + NC-t2t}.  
Similar patterns are also reproduced while using $ D\textsuperscript{-}(I;C) $.
These findings highlight the effectiveness of the proposed methodology. By producing ``harder'' training samples and reducing the rate of \AMM, \methodname \ can significantly improve predictive performance on real-world data.
Finally, it is worth noting that while $D(\cdot)$ trained on \methodname \ displayed a high rate of text-side unimodal bias on COSMOS, this phenomenon is not present in the \benchmark \ evaluation benchmark.

\section{Conclusions}

In this study, we address the task of multimodal misinformation detection (\MMD) where an image and its accompanying caption collaborate to spread misleading or false information. 
Our primary focus lies in addressing the issue of unimodal bias, arising in datasets that exhibit distinct patterns and biases towards one modality, which allows unimodal methods to outperform their multimodal counterparts in an inherently multimodal task.
Our systematic investigation found that datasets widely used for \MMD, namely \twitterdataset \ and COSMOS, can enable image-side and text-side unimodal bias respectively; raising questions about their reliability as benchmarks for \MMD. 

To address the aforementioned concerns, we introduce the \benchmark \ evaluation benchmark, designed to provide a comprehensive and robust framework for multimodal misinformation detection. 
\benchmark \ encompasses a diverse array of real-world data, excludes ``asymmetric multimodal misinformation'' (\AMM) -where one modality plays a dominant role in propagating misinformation while others have little or no influence- and implements ``modality balancing''; where each image and caption appear twice in the dataset, once within their truthful and once within a misleading pair.
We conduct an extensive comparative study with a Transformer-based architecture which demonstrates that \benchmark \ effectively mitigates and prevents the manifestation of unimodal bias, offering an improved evaluation framework for \MMD.

In addition, we introduce \textit{\methodname}, a novel method for generating synthetic training data  
that retain crossmodal relations between image-caption pairs. \textit{\methodname} employs a large pre-trained crossmodal alignment model to generate hard examples that retain crossmodal relations between legitimate images and misleading human-written captions.
Empirical results show that using \textit{\methodname} in the training process consistently improves detection accuracy and has achieved the highest performance on \benchmark. 

The proposed approach achieved 52.1\% accuracy for multiclass \MMD. 
Nevertheless, we are optimistic that \methodname \ and \benchmark \ can serve as a foundation for future research, leading to further advancements in this area.
For instance, future works could experiment with improved multimodal encoders \cite{li2021align, li2023blip}, news- or event-aware encoders \cite{li2022clip},
advanced modality fusion techniques \cite{yu2022bcmf, wu2021multimodal, jing2023multimodal}, utilize external evidence \cite{abdelnabi2022open} or explore new methods for generating training data \cite{cardenuto2023age}.
As future research unfolds, \benchmark \ could be expanded to include additional types of MM (e.g. AI generated content) or additional modalities (e.g. videos), or be repurposed for other relevant tasks (e.g. fact-checked article retrieval \cite{nakov2021clef}). Moreover, since \MMD \ is only one part of multimedia verification \cite{mubashara2023multimodal}, ``claim detection'' and ``check-worthiness'' \cite{cheema2022mm} could be employed to distinguish between \AMM \ and MM and determine whether to use a unimodal detector (e.g. false claim or manipulated image) or a multimodal misinformation detector in each scenario.
Finally, while our focus has been on alleviating unimodal bias at the evaluation level, it may be worth exploring methods for reducing unimodal bias from an algorithmic perspective \cite{cadene2019rubi}.
In all these endeavors, \benchmark \ can serve as a robust evaluation benchmark.

\section{Acknowledgments}

This work is partially funded by the project ``vera.ai: VERification Assisted by Artificial Intelligence'' under grant agreement no. 101070093.

\bibliographystyle{unsrt}  
\bibliography{references}

\begin{thebibliography}{10}

\bibitem{bennett2018disinformation}
W~Lance Bennett and Steven Livingston.
\newblock The disinformation order: Disruptive communication and the decline of
  democratic institutions.
\newblock {\em European journal of communication}, 33(2):122--139, 2018.

\bibitem{duffy2020too}
Andrew Duffy, Edson Tandoc, and Rich Ling.
\newblock Too good to be true, too good not to share: the social utility of
  fake news.
\newblock {\em Information, Communication \& Society}, 23(13):1965--1979, 2020.

\bibitem{roozenbeek2020susceptibility}
Jon Roozenbeek, Claudia~R Schneider, Sarah Dryhurst, John Kerr, Alexandra~LJ
  Freeman, Gabriel Recchia, Anne~Marthe Van Der~Bles, and Sander Van
  Der~Linden.
\newblock Susceptibility to misinformation about covid-19 around the world.
\newblock {\em Royal Society open science}, 7(10):201199, 2020.

\bibitem{gamir2021multimodal}
Jos{\'e} Gamir-R{\'\i}os, Raquel Tarullo, Miguel Ib{\'a}{\~n}ez-Cuquerella,
  et~al.
\newblock Multimodal disinformation about otherness on the internet. the spread
  of racist, xenophobic and islamophobic fake news in 2020.
\newblock {\em An{\`a}lisi}, pages 49--64, 2021.

\bibitem{olan2022fake}
Femi Olan, Uchitha Jayawickrama, Emmanuel~Ogiemwonyi Arakpogun, Jana Suklan,
  and Shaofeng Liu.
\newblock Fake news on social media: the impact on society.
\newblock {\em Information Systems Frontiers}, pages 1--16, 2022.

\bibitem{li2020picture}
Yiyi Li and Ying Xie.
\newblock Is a picture worth a thousand words? an empirical study of image
  content and social media engagement.
\newblock {\em Journal of Marketing Research}, 57(1):1--19, 2020.

\bibitem{newman2012nonprobative}
Eryn~J Newman, Maryanne Garry, Daniel~M Bernstein, Justin Kantner, and
  D~Stephen Lindsay.
\newblock Nonprobative photographs (or words) inflate truthiness.
\newblock {\em Psychonomic Bulletin \& Review}, 19:969--974, 2012.

\bibitem{mridha2021comprehensive}
Muhammad~F Mridha, Ashfia~Jannat Keya, Md~Abdul Hamid, Muhammad~Mostafa
  Monowar, and Md~Saifur Rahman.
\newblock A comprehensive review on fake news detection with deep learning.
\newblock {\em IEEE Access}, 9:156151--156170, 2021.

\bibitem{rana2022deepfake}
Md~Shohel Rana, Mohammad~Nur Nobi, Beddhu Murali, and Andrew~H Sung.
\newblock Deepfake detection: A systematic literature review.
\newblock {\em IEEE Access}, 2022.

\bibitem{hangloo2022combating}
Sakshini Hangloo and Bhavna Arora.
\newblock Combating multimodal fake news on social media: methods, datasets,
  and future perspective.
\newblock {\em Multimedia Systems}, 28(6):2391--2422, 2022.

\bibitem{alam2021survey}
Firoj Alam, Stefano Cresci, Tanmoy Chakraborty, Fabrizio Silvestri, Dimiter
  Dimitrov, Giovanni Da~San Martino, Shaden Shaar, Hamed Firooz, and Preslav
  Nakov.
\newblock A survey on multimodal disinformation detection.
\newblock {\em arXiv preprint arXiv:2103.12541}, 2021.

\bibitem{boididou2018verifying}
Christina Boididou, Stuart~E Middleton, Zhiwei Jin, Symeon Papadopoulos,
  Duc-Tien Dang-Nguyen, Giulia Boato, and Yiannis Kompatsiaris.
\newblock Verifying information with multimedia content on twitter: a
  comparative study of automated approaches.
\newblock {\em Multimedia tools and applications}, 77:15545--15571, 2018.

\bibitem{zlatkova2019fact}
Dimitrina Zlatkova, Preslav Nakov, and Ivan Koychev.
\newblock Fact-checking meets fauxtography: Verifying claims about images.
\newblock {\em arXiv preprint arXiv:1908.11722}, 2019.

\bibitem{nielsen2022mumin}
Dan~S Nielsen and Ryan McConville.
\newblock Mumin: A large-scale multilingual multimodal fact-checked
  misinformation social network dataset.
\newblock In {\em Proceedings of the 45th International ACM SIGIR Conference on
  Research and Development in Information Retrieval}, pages 3141--3153, 2022.

\bibitem{jindal2020newsbag}
Sarthak Jindal, Raghav Sood, Richa Singh, Mayank Vatsa, and Tanmoy Chakraborty.
\newblock Newsbag: A multimodal benchmark dataset for fake news detection.
\newblock In {\em CEUR Workshop Proc.}, volume 2560, pages 138--145, 2020.

\bibitem{nakamura2019r}
Kai Nakamura, Sharon Levy, and William~Yang Wang.
\newblock r/fakeddit: A new multimodal benchmark dataset for fine-grained fake
  news detection.
\newblock {\em arXiv preprint arXiv:1911.03854}, 2019.

\bibitem{jaiswal2017multimedia}
Ayush Jaiswal, Ekraam Sabir, Wael AbdAlmageed, and Premkumar Natarajan.
\newblock Multimedia semantic integrity assessment using joint embedding of
  images and text.
\newblock In {\em Proceedings of the 25th ACM international conference on
  Multimedia}, pages 1465--1471, 2017.

\bibitem{luo2021newsclippings}
Grace Luo, Trevor Darrell, and Anna Rohrbach.
\newblock Newsclippings: Automatic generation of out-of-context multimodal
  media.
\newblock {\em arXiv preprint arXiv:2104.05893}, 2021.

\bibitem{sabir2018deep}
Ekraam Sabir, Wael AbdAlmageed, Yue Wu, and Prem Natarajan.
\newblock Deep multimodal image-repurposing detection.
\newblock In {\em Proceedings of the 26th ACM international conference on
  Multimedia}, pages 1337--1345, 2018.

\bibitem{wang2018eann}
Yaqing Wang, Fenglong Ma, Zhiwei Jin, Ye~Yuan, Guangxu Xun, Kishlay Jha, Lu~Su,
  and Jing Gao.
\newblock Eann: Event adversarial neural networks for multi-modal fake news
  detection.
\newblock In {\em Proceedings of the 24th acm sigkdd international conference
  on knowledge discovery \& data mining}, pages 849--857, 2018.

\bibitem{khattar2019mvae}
Dhruv Khattar, Jaipal~Singh Goud, Manish Gupta, and Vasudeva Varma.
\newblock Mvae: Multimodal variational autoencoder for fake news detection.
\newblock In {\em The world wide web conference}, pages 2915--2921, 2019.

\bibitem{singhal2019spotfake}
Shivangi Singhal, Rajiv~Ratn Shah, Tanmoy Chakraborty, Ponnurangam Kumaraguru,
  and Shin'ichi Satoh.
\newblock Spotfake: A multi-modal framework for fake news detection.
\newblock In {\em 2019 IEEE fifth international conference on multimedia big
  data (BigMM)}, pages 39--47. IEEE, 2019.

\bibitem{yu2022bcmf}
Chuanming Yu, Yinxue Ma, Lu~An, and Gang Li.
\newblock Bcmf: A bidirectional cross-modal fusion model for fake news
  detection.
\newblock {\em Information Processing \& Management}, 59(5):103063, 2022.

\bibitem{aneja2021cosmos}
Shivangi Aneja, Chris Bregler, and Matthias Nie{\ss}ner.
\newblock Cosmos: Catching out-of-context misinformation with self-supervised
  learning.
\newblock {\em arXiv preprint arXiv:2101.06278}, 2021.

\bibitem{radford2021learning}
Alec Radford, Jong~Wook Kim, Chris Hallacy, Aditya Ramesh, Gabriel Goh,
  Sandhini Agarwal, Girish Sastry, Amanda Askell, Pamela Mishkin, Jack Clark,
  et~al.
\newblock Learning transferable visual models from natural language
  supervision.
\newblock In {\em International conference on machine learning}, pages
  8748--8763. PMLR, 2021.

\bibitem{liu2020visual}
Fuxiao Liu, Yinghan Wang, Tianlu Wang, and Vicente Ordonez.
\newblock Visual news: Benchmark and challenges in news image captioning.
\newblock {\em arXiv preprint arXiv:2010.03743}, 2020.

\bibitem{thorne2018fever}
James Thorne, Andreas Vlachos, Christos Christodoulopoulos, and Arpit Mittal.
\newblock Fever: a large-scale dataset for fact extraction and verification.
\newblock {\em arXiv preprint arXiv:1803.05355}, 2018.

\bibitem{heller2018ps}
Silvan Heller, Luca Rossetto, and Heiko Schuldt.
\newblock The ps-battles dataset-an image collection for image manipulation
  detection.
\newblock {\em arXiv preprint arXiv:1804.04866}, 2018.

\bibitem{levi2019identifying}
Or~Levi, Pedram Hosseini, Mona Diab, and David~A Broniatowski.
\newblock Identifying nuances in fake news vs. satire: using semantic and
  linguistic cues.
\newblock {\em arXiv preprint arXiv:1910.01160}, 2019.

\bibitem{papadopoulos2023synthetic}
Stefanos-Iordanis Papadopoulos, Christos Koutlis, Symeon Papadopoulos, and
  Panagiotis Petrantonakis.
\newblock Synthetic misinformers: Generating and combating multimodal
  misinformation.
\newblock In {\em Proceedings of the 2nd ACM International Workshop on
  Multimedia AI against Disinformation}, pages 36--44, 2023.

\bibitem{biamby2021twitter}
Giscard Biamby, Grace Luo, Trevor Darrell, and Anna Rohrbach.
\newblock Twitter-comms: Detecting climate, covid, and military multimodal
  misinformation.
\newblock {\em arXiv preprint arXiv:2112.08594}, 2021.

\bibitem{muller2020multimodal}
Eric M{\"u}ller-Budack, Jonas Theiner, Sebastian Diering, Maximilian Idahl, and
  Ralph Ewerth.
\newblock Multimodal analytics for real-world news using measures of
  cross-modal entity consistency.
\newblock In {\em Proceedings of the 2020 International Conference on
  Multimedia Retrieval}, pages 16--25, 2020.

\bibitem{goyal2017making}
Yash Goyal, Tejas Khot, Douglas Summers-Stay, Dhruv Batra, and Devi Parikh.
\newblock Making the v in vqa matter: Elevating the role of image understanding
  in visual question answering.
\newblock In {\em Proceedings of the IEEE conference on computer vision and
  pattern recognition}, pages 6904--6913, 2017.

\bibitem{agrawal2018don}
Aishwarya Agrawal, Dhruv Batra, Devi Parikh, and Aniruddha Kembhavi.
\newblock Don't just assume; look and answer: Overcoming priors for visual
  question answering.
\newblock In {\em Proceedings of the IEEE conference on computer vision and
  pattern recognition}, pages 4971--4980, 2018.

\bibitem{cadene2019rubi}
Remi Cadene, Corentin Dancette, Matthieu Cord, Devi Parikh, et~al.
\newblock Rubi: Reducing unimodal biases for visual question answering.
\newblock {\em Advances in neural information processing systems}, 32, 2019.

\bibitem{aneja2021mmsys}
Shivangi Aneja, Cise Midoglu, Duc-Tien Dang-Nguyen, Michael~Alexander Riegler,
  Paal Halvorsen, Matthias Nie{\ss}ner, Balu Adsumilli, and Chris Bregler.
\newblock Mmsys' 21 grand challenge on detecting cheapfakes.
\newblock {\em arXiv preprint arXiv:2107.05297}, 2021.

\bibitem{aneja2022acm}
Shivangi Aneja, Cise Midoglu, Duc-Tien Dang-Nguyen, Sohail~Ahmed Khan, Michael
  Riegler, P{\aa}l Halvorsen, Chris Bregler, and Balu Adsumilli.
\newblock Acm multimedia grand challenge on detecting cheapfakes.
\newblock {\em arXiv preprint arXiv:2207.14534}, 2022.

\bibitem{lin2022frozen}
Ziyi Lin, Shijie Geng, Renrui Zhang, Peng Gao, Gerard de~Melo, Xiaogang Wang,
  Jifeng Dai, Yu~Qiao, and Hongsheng Li.
\newblock Frozen clip models are efficient video learners.
\newblock In {\em European Conference on Computer Vision}, pages 388--404.
  Springer, 2022.

\bibitem{guzhov2022audioclip}
Andrey Guzhov, Federico Raue, J{\"o}rn Hees, and Andreas Dengel.
\newblock Audioclip: Extending clip to image, text and audio.
\newblock In {\em ICASSP 2022-2022 IEEE International Conference on Acoustics,
  Speech and Signal Processing (ICASSP)}, pages 976--980. IEEE, 2022.

\bibitem{li2022clip}
Manling Li, Ruochen Xu, Shuohang Wang, Luowei Zhou, Xudong Lin, Chenguang Zhu,
  Michael Zeng, Heng Ji, and Shih-Fu Chang.
\newblock Clip-event: Connecting text and images with event structures.
\newblock In {\em Proceedings of the IEEE/CVF Conference on Computer Vision and
  Pattern Recognition}, pages 16420--16429, 2022.

\bibitem{zhang2023ino}
Yinuo Zhang, Zhulin Tao, Xi~Wang, and Tongyue Wang.
\newblock Ino at factify 2: Structure coherence based multi-modal fact
  verification.
\newblock {\em arXiv preprint arXiv:2303.01510}, 2023.

\bibitem{zhou2023multimodal}
Yangming Zhou, Yuzhou Yang, Qichao Ying, Zhenxing Qian, and Xinpeng Zhang.
\newblock Multimodal fake news detection via clip-guided learning.
\newblock In {\em 2023 IEEE International Conference on Multimedia and Expo
  (ICME)}, pages 2825--2830. IEEE, 2023.

\bibitem{tahmasebi2023improving}
Sahar Tahmasebi, Sherzod Hakimov, Ralph Ewerth, and Eric M{\"u}ller-Budack.
\newblock Improving generalization for multimodal fake news detection.
\newblock {\em arXiv preprint arXiv:2305.18599}, 2023.

\bibitem{vaswani2017attention}
Ashish Vaswani, Noam Shazeer, Niki Parmar, Jakob Uszkoreit, Llion Jones,
  Aidan~N Gomez, {\L}ukasz Kaiser, and Illia Polosukhin.
\newblock Attention is all you need.
\newblock {\em Advances in neural information processing systems}, 30, 2017.

\bibitem{singhal2022leveraging}
Shivangi Singhal, Tanisha Pandey, Saksham Mrig, Rajiv~Ratn Shah, and
  Ponnurangam Kumaraguru.
\newblock Leveraging intra and inter modality relationship for multimodal fake
  news detection.
\newblock In {\em Companion Proceedings of the Web Conference 2022}, pages
  726--734, 2022.

\bibitem{koh2021wilds}
Pang~Wei Koh, Shiori Sagawa, Henrik Marklund, Sang~Michael Xie, Marvin Zhang,
  Akshay Balsubramani, Weihua Hu, Michihiro Yasunaga, Richard~Lanas Phillips,
  Irena Gao, et~al.
\newblock Wilds: A benchmark of in-the-wild distribution shifts.
\newblock In {\em International Conference on Machine Learning}, pages
  5637--5664. PMLR, 2021.

\bibitem{li2021align}
Junnan Li, Ramprasaath Selvaraju, Akhilesh Gotmare, Shafiq Joty, Caiming Xiong,
  and Steven Chu~Hong Hoi.
\newblock Align before fuse: Vision and language representation learning with
  momentum distillation.
\newblock {\em Advances in neural information processing systems},
  34:9694--9705, 2021.

\bibitem{li2023blip}
Junnan Li, Dongxu Li, Silvio Savarese, and Steven Hoi.
\newblock Blip-2: Bootstrapping language-image pre-training with frozen image
  encoders and large language models.
\newblock {\em arXiv preprint arXiv:2301.12597}, 2023.

\bibitem{wu2021multimodal}
Yang Wu, Pengwei Zhan, Yunjian Zhang, Liming Wang, and Zhen Xu.
\newblock Multimodal fusion with co-attention networks for fake news detection.
\newblock In {\em Findings of the association for computational linguistics:
  ACL-IJCNLP 2021}, pages 2560--2569, 2021.

\bibitem{jing2023multimodal}
Jing Jing, Hongchen Wu, Jie Sun, Xiaochang Fang, and Huaxiang Zhang.
\newblock Multimodal fake news detection via progressive fusion networks.
\newblock {\em Information processing \& management}, 60(1):103120, 2023.

\bibitem{abdelnabi2022open}
Sahar Abdelnabi, Rakibul Hasan, and Mario Fritz.
\newblock Open-domain, content-based, multi-modal fact-checking of
  out-of-context images via online resources.
\newblock In {\em Proceedings of the IEEE/CVF Conference on Computer Vision and
  Pattern Recognition}, pages 14940--14949, 2022.

\bibitem{cardenuto2023age}
Jo{\~a}o~Phillipe Cardenuto, Jing Yang, Rafael Padilha, Renjie Wan, Daniel
  Moreira, Haoliang Li, Shiqi Wang, Fernanda Andal{\'o}, S{\'e}bastien Marcel,
  and Anderson Rocha.
\newblock The age of synthetic realities: Challenges and opportunities.
\newblock {\em arXiv preprint arXiv:2306.11503}, 2023.

\bibitem{nakov2021clef}
Preslav Nakov, Giovanni Da~San~Martino, Tamer Elsayed, Alberto
  Barr{\'o}n-Cedeno, Rub{\'e}n M{\'\i}guez, Shaden Shaar, Firoj Alam, Fatima
  Haouari, Maram Hasanain, Nikolay Babulkov, et~al.
\newblock The clef-2021 checkthat! lab on detecting check-worthy claims,
  previously fact-checked claims, and fake news.
\newblock In {\em Advances in Information Retrieval: 43rd European Conference
  on IR Research, ECIR 2021, Virtual Event, March 28--April 1, 2021,
  Proceedings, Part II 43}, pages 639--649. Springer, 2021.

\bibitem{mubashara2023multimodal}
Akhtar Mubashara, Schlichtkrull Michael, Guo Zhijiang, Cocarascu Oana, Simperl
  Elena, and Vlachos Andreas.
\newblock Multimodal automated fact-checking: A survey.
\newblock {\em arXiv preprint arXiv:2305.13507}, 2023.

\bibitem{cheema2022mm}
Gullal~S Cheema, Sherzod Hakimov, Abdul Sittar, Eric M{\"u}ller-Budack,
  Christian Otto, and Ralph Ewerth.
\newblock Mm-claims: a dataset for multimodal claim detection in social media.
\newblock {\em arXiv preprint arXiv:2205.01989}, 2022.

\end{thebibliography}

\end{document}